%% file: main.tex
\documentclass[10pt,twocolumn,letterpaper]{article}

\usepackage[pagenumbers]{iccv} 
\usepackage{mathtools}
\usepackage{graphicx}
\usepackage{booktabs}
\usepackage{amssymb}
\usepackage{amsmath}
\usepackage{pifont}
\newcommand{\cmark}{\ding{51}}%
\newcommand{\xmark}{\ding{55}}%
\usepackage{arydshln}
\usepackage{tabularx}
\usepackage{float}
\usepackage[accsupp]{axessibility}  
\usepackage[dvipsnames]{xcolor}


\definecolor{iccvblue}{rgb}{0.21,0.49,0.74}
\usepackage[pagebackref,breaklinks,colorlinks,allcolors=iccvblue]{hyperref}

\def\paperID{9090} 
\def\confName{ICCV}
\def\confYear{2025}

\title{Clink! Chop! Thud! -- Learning Object Sounds from Real-World Interactions}

\author{Mengyu Yang$^{1}$ \hspace{5mm} Yiming Chen$^{1}$ \hspace{5mm} Haozheng Pei$^{1}$ \hspace{5mm} Siddhant Agarwal$^{1}$\\
 Arun Balajee Vasudevan$^{2}$ \hspace{5mm} James Hays$^{1}$\\
\small{$^{1}$Georgia Institute of Technology \hspace{5mm} $^{2}$Carnegie Mellon University}\\
\small{Project Page: \url{https://clink-chop-thud.github.io/}}
}

\begin{document}

\input{sec/_teaser}
\input{sec/0_abstract}    
\input{sec/1_intro}
\input{sec/2_relatedworks}
\input{sec/3_tasks}
\input{sec/4_method}

\input{sec/5_datasets}
\input{sec/6_experiments}

\input{sec/7_conclusion}

\clearpage

{
    \small
    \bibliographystyle{ieeenat_fullname}
    \bibliography{main}
}

\clearpage

\input{sec/8_supplementary}

\end{document}

%% file: sec/_teaser.tex
\twocolumn[{%
\renewcommand\twocolumn[1][]{#1}%
\maketitle
\vspace{-2em}
\centering
\includegraphics[width=0.9\linewidth]{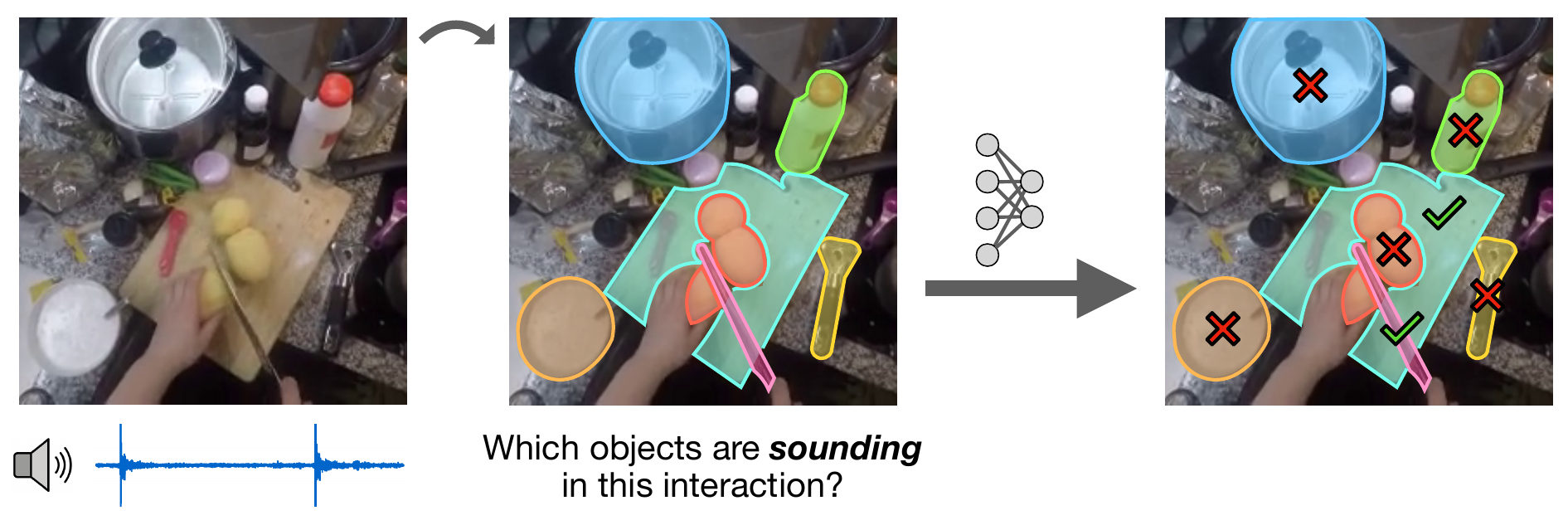}
\captionof{figure}{Humans handle a wide variety of objects throughout the day and many of these interactions produce sounds. We introduce a multimodal \textit{object-aware} framework that learns the relationship between the objects in an interaction and the resulting sounds. This enables our model to detect the \textit{sounding objects} from a set of candidates in a scene.
\vspace{2em}}
\label{fig:teaser}
}]

%% file: sec/0_abstract.tex
\begin{abstract}
Can a model distinguish between the sound of a spoon hitting a hardwood floor versus a carpeted one? Everyday object interactions produce sounds unique to the objects involved. We introduce the sounding object detection task to evaluate a model's ability to link these sounds to the objects directly involved. Inspired by human perception, our multimodal object-aware framework learns from in-the-wild egocentric videos. To encourage an object-centric approach, we first develop an automatic pipeline to compute segmentation masks of the objects involved to guide the model's focus during training towards the most informative regions of the interaction. A slot attention visual encoder is used to further enforce an object prior. We demonstrate state of the art performance on our new task along with existing multimodal action understanding tasks.\vspace{-4mm}
\end{abstract}

%% file: sec/1_intro.tex
\section{Introduction}
\label{sec:intro}

Humans interact with many physical objects throughout the day. Oftentimes, interacting with an object results in a distinct sound that provides informative cues about that action and the objects involved. The sound produced when a knife cuts through an onion is distinct from the sound made when the knife hits the chopping board. Humans can listen to the unique sounds of these object interactions and make cross-modal inferences about the objects involved. After hearing a sound, we can look at a visual scene and latch onto its \textit{focal region}. For example, if we hear the sound of oil sizzling, we do not first look at the sink next to the stove but instead direct our attention to the person stirring food in a pan. This is a biological phenomenon where humans perceive their surroundings by concentrating on key regions in a scene~\cite{Grill-Spector2005-or,tenenbaum2011how}. While humans excel at distinguishing these sounds, current learning methods face challenges. Unlike the typical video datasets used in multimodal training~\cite{aytar2016soundnet, chen2020vggsound, gemmeke2017audioset}, which feature predefined sound categories (e.g. vehicles, animals), differentiating the impact sound caused by different materials is more subtle. This shortcoming is compounded by vision frameworks relying on global features that take the whole scene as input~\cite{arnab2021vivit,bertasius2021spacetime,liu2021videoswin,wang2022adafocusv3}. Meanwhile, understanding object interactions requires a localized, object-centric approach.

We evaluate a model's understanding of the relationship between object interactions and sound by proposing the sounding object detection task. Given a set of object regions present in a scene and the sound of an interaction, the model predicts which object is directly involved in producing the sound. To test this, we introduce a manually annotated benchmark containing segmentation masks of the ground truth objects. We further evaluate the model's understanding of sounds and object interactions by having it predict whether the existing sound is caused by the interaction in the first place. This task, called sounding action discovery, was first proposed by Chen \etal~\cite{chen2024soundingactions}. We demonstrate that with our method, we outperform existing multimodal techniques on both tasks.

We do so by taking inspiration from natural human perception and develop a multimodal \textit{object-aware} framework to guide our model to explicitly focus on the informative regions of an object interaction. We leverage in-the-wild egocentric videos of interactions from Ego4D~\cite{grauman2022ego4d} and Epic Kitchens~\cite{Damen2021PAMI}, with the benefits being twofold: 1) The large scale of these datasets means that a wide range of objects and object interactions are captured, including long-tail examples, and 2) having an egocentric point of view means most actions occur in the near field, making the objects more visible and the interactions more audible.

While existing works have studied frameworks that focus on objects to reduce visual redundancy and learn a more compact representation~\cite{zhou2023objects,locatello2020object,shamsian2020learning,elsayed2022savi}, these works only learn from a single modality and exclude the additional information gained by associating object interactions with sounds. There are a few works on multimodal object-centric representation learning~\cite{gao2021ObjectFolder, gao2022ObjectFolderV2} but the data they train on is synthetic or laboratory-collected. Such methods do not scale well if we want to learn about the wide range of object interactions that occur in daily life.

To help our method scale to these larger datasets, we develop an automatic pipeline that leverages an off-the-shelf hand-object interaction detection model~\cite{Shan2020understanding} to automatically annotate our training data with object segmentation masks. These masks inform our model on which visual patch features to sample, guiding its focus towards the regions of the object interaction and learning an object-aware representation in the process. To further equip our model with a strong object prior, we initialize our visual encoder with a pretrained slot attention model~\cite{Kakogeorgiou2024SPOT}. These architectures contain a bottleneck attention module which learn to compress visual features into $k$ output vectors called slots~\cite{locatello2020object}. Through training, each slot learns to attend to unique objects and creates object-centric boundaries. Using our multimodal object-aware approach, we aim to learn the unique correlations between an object interaction and the sound it produces. To do so, we introduce the sounding object detection task along with a manually annotated benchmark and demonstrate state of the art performance.

%% file: sec/2_relatedworks.tex
\section{Related Work}
\label{sec:_relatedworks}

\paragraph{Multimodal representation learning} 

Multimodal representation learning unifies vision, audio, language, and other modalities through contrastive and self-supervised learning. Contrastive learning on large-scale datasets have enabled vision-language pretraining for zero-shot classification and retrieval, later extending to audio-text alignment~\cite{radford2021learning, wu2023large}. In ImageBind~\cite{girdhar2023imagebind}, a unified embedding space is learned across a range of modalities to enable cross-modal associations without direct supervision. This was later refined by mapping all modalities to language for stronger semantic consistency~\cite{zhu2023languagebind}. In other works,~\cite{zhang2023meta} supports twelve modalities with different tokenization techniques using a scalable multimodal framework. In~\cite{chen2024soundingactions}, the authors develop a contrastive-consensus loss framework to learn how actions sound from egocentric videos. The loss refines audio-vision-language embeddings to better associate actions with their characteristic sounds. Our work relies on similar contrastive frameworks to learn from egocentric videos, but we employ an object-aware approach that enables our model to reason about localized object interactions.

\paragraph{Object representations} 

Traditional deep learning models typically analyze entire scenes without distinguishing between individual objects, leading to inefficiencies which object-centric learning aims to address. In~\cite{zhou2023can}, object-guided token sampling and attention are used to enhance action recognition in video transformers. Additionally, slot attention models~\cite{locatello2020object} use a bottleneck attention module to compress image features into $k$ slot vectors. Through an iterative attention mechanism, slots learn to attend to individual object regions, forming object segmentation masks.~\cite{Kakogeorgiou2024SPOT} introduced improvements to the original slot attention model by using a student-teacher framework. Finally,~\cite{gao2021ObjectFolder, gao2022ObjectFolderV2} introduced a multimodal dataset to capture high-fidelity object properties across RGB, depth, touch, and audio, enabling object-centric learning beyond visual cues. However, the data is generated synthetically or collected in a controlled lab environment. On the other hand, our work learns from in-the-wild egocentric videos which capture a wide range of objects and object interactions.

\paragraph{Audio-visual correspondence} 

Natural correlations exist between vision and audio and jointly modeling them can lead to richer representations~\cite{arandjelovic2017look, afouras2020self,vasudevan2022sound,chen2024soundingactions}. Tasks that leverage this correlation include using vision as supervision for sounds~\cite{aytar2016soundnet} and using the temporal~\cite{korbar2018cooperative} and spatial~\cite{vasudevan2022sound} alignment as a self-supervisory signal. Additional lines of work involve tasks such as sound source localization~\cite{gan2019self,yang2024kidnappable,vasudevan2020semantic} and audiovisual localization~\cite{mo2022localizing, chen2021localizing, mo2022SLAVC}. The former involves using vision as a supervisory signal, such as when~\cite{yang2024kidnappable} captured the subtle sounds of human movement and learned to ground the sound source to the humans in the video, while the latter learns to semantically associate visual regions with audio representations. Our work focuses on explicitly associating in-the-wild objects with the sounds of its interactions, where the differences can be subtle and less semantically obvious.


%% file: sec/3_tasks.tex
\section{Task formulation}
\label{sec:_tasks}

\paragraph{Sounding object detection}

We define a \textit{sounding object} to be an object that is directly involved in an action, where the resulting interaction generates sound. While an environment may contain many objects, as shown in \cref{fig:teaser}, only one or two will be central to the action. We propose to detect these salient objects by leveraging the sound produced during the interaction, as it can reveal object contact and distinguish between characteristics like material and mass.

Given a set of object regions in a scene along with the video and sound of an object interaction, we want to detect the objects that are directly involved in the action. While this task seems similar to audiovisual localization~\cite{senocak2018learning, mo2022localizing, chen2021localizing}, the latter is more under-constrained as it requires the model itself to predict the boundaries of a sounding region. However, our work is more interested in \textit{which} objects are involved rather than \textit{where} exactly that object is. And given our object-centric framework where we assume the presence of objects, it is easier for us to predetermine the boundaries of all potential objects and have our model make a decision based on that information. Thus, our model is more focused on predicting the semantic relationships between distinct objects and sounds rather than their visual boundaries, which is less relevant to our work. 

Along the same lines, the typical metrics associated with audiovisual localization such as consensus intersection over union~\cite{senocak2018learning} penalize predictions that do not exactly align with the ground truth boundaries. Even if a localization map mainly centers its prediction on the correct object, some stray predictions will lead to a lower cIoU score. Whereas in our sounding object detection task, models predict a similarity score for each given object and we use top-1 accuracy to measure performance. Our metric puts the focus on the comparison between different objects rather than the exact boundaries of a specific object.

\paragraph{Sounding action discovery}

\begin{figure}
  \centering
   \includegraphics[width=.95\columnwidth]{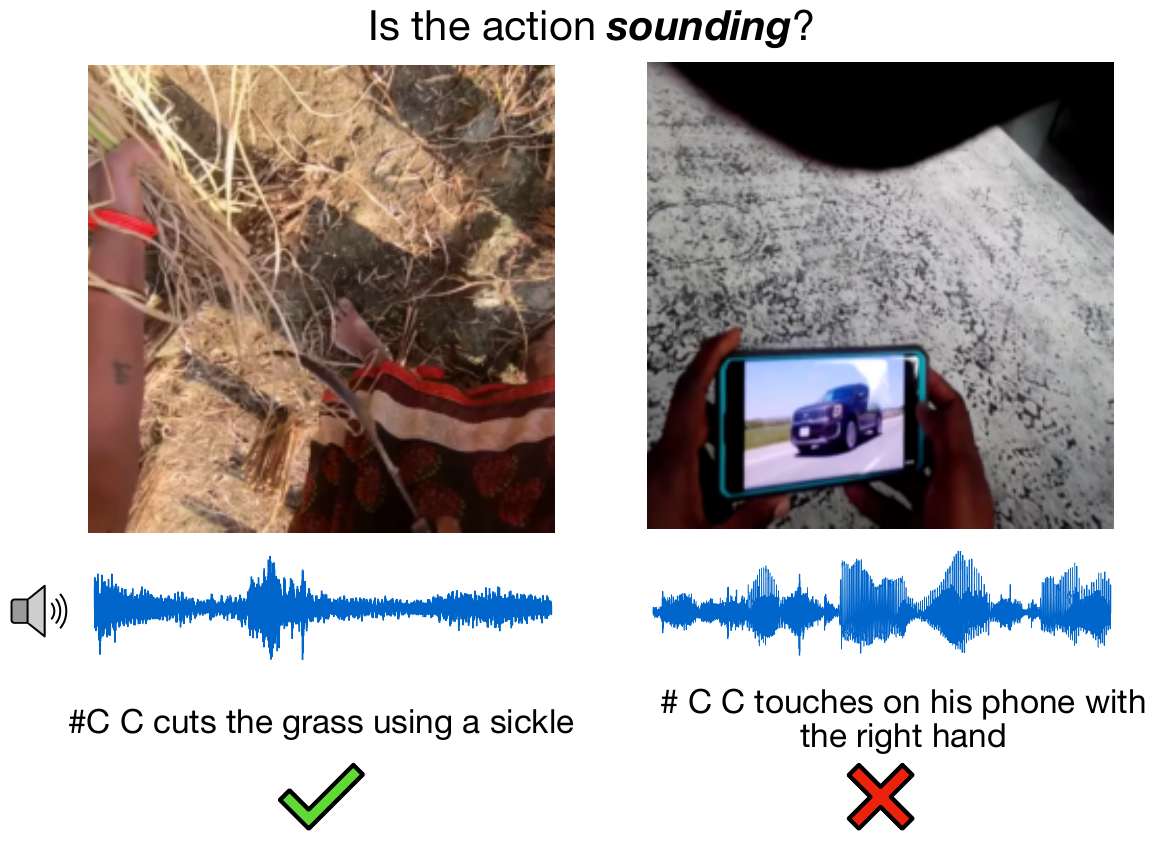}
   \caption{We also evaluate our model on sounding action discovery~\cite{chen2024soundingactions}. The left example shows a sounding action, where cutting the grass directly produces the rustling sound. Meanwhile, the right example depicts a non-sounding action where the sound comes from the video and not the action of tapping the screen.}
   \label{fig:sounding_actions}
\end{figure}

Closely tied to the concept of sounding objects are \textit{sounding actions}, first introduced by Chen \etal~\cite{chen2024soundingactions}. They define a sounding action as ``a human-initiated action that produces sound during its execution due to interactions with the surrounding environment''. Based on this definition, one can think of sounding objects as the specific elements in that environment. Examples of sounding actions are shown in \cref{fig:sounding_actions}.

This task includes language as a third modality, where the narrations found in Ego4D~\cite{grauman2022ego4d} and Epic Kitchens~\cite{Damen2021PAMI} directly describe the action depicted. As a result, we can leverage these narrations as the ground truth to see if both the visual and audio modalities align. If, for example, the video depicts the narrated action but the audio is dominated by off-screen sounds like music, then this mismatch indicates that the action is not sounding. This alignment is leveraged when the authors design their loss, detailed in \cref{sec:training_framework}. While sounding object detection requires a localized understanding of a scene, sounding action discovery is a more global task. Given a video of an action and the corresponding sound, a model performs binary classification to predict whether the action is sounding or not.

\paragraph{Putting it together}

We consider sounding object detection and sounding action discovery as complementary and evaluate our method on both. Having a model capable of both tasks allows us to first determine which actions are sounding and then among the sounding actions, we can detect the exact objects that caused the sound. Such a framework allows us to understand the relationship between visual actions and sounds at both a global and localized, object-centric level.

%% file: sec/4_method.tex
\section{Method}
\label{sec:_methods}

\begin{figure*}[ht]
  \centering
   \includegraphics[width=0.9\linewidth]{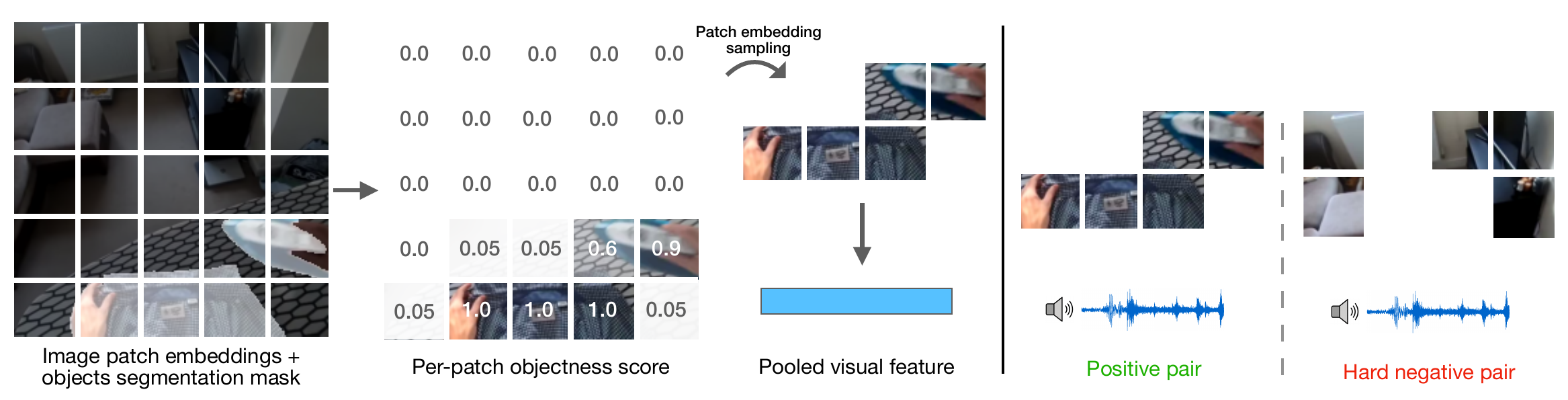}
   \caption{\textbf{Left}: Our object-aware visual features. Given a video frame and corresponding objects segmentation mask, we first encode the image into patch embeddings. We also patchify the mask to get a per-patch objectness score, corresponding to the percentage of the patch containing the object. The score informs the model on which patch embeddings to keep based on a threshold and the remaining embeddings are pooled into a single visual embedding vector. \textbf{Right}: The hard negatives paradigm used in the finetuning stage. Additional negative embeddings are sampled from non-interaction regions of the same image.}
   \label{fig:visual_framework}
\end{figure*}

We introduce a framework to learn about object interactions across vision, audio, and language. Our model learns object-aware representations by focusing on the regions where the interaction occurs. We train our model contrastively to align all modalities into a common embedding space. By doing so, we can directly compute the similarity between modality embeddings using cosine similarity.

\subsection{Architecture} \label{sec:architecture}

Formally, our model takes as input video frames $V \in \mathbb{R}^{T \times H \times W \times 3}$, audio waveform $A \in \mathbb{R}^{S}$, and language narration $L$, where $T$ is the number of video frames and $S$ is the sequence length of the audio waveform. Each modality has its own encoder which converts the raw input into learned embeddings $e_v^* \in \mathbb{R}^{T \times N \times D}$, $e_a \in \mathbb{R}^{D}$, and $e_l \in \mathbb{R}^{D}$, respectively. Here, $N$ is the number of non-overlapping patches per frame and $D$ is the common feature dimension which all modalities are projected to. To make our model object-aware, we also have binary object segmentation masks of the objects involved in the interaction $M \in \mathbb{R}^{T \times H \times W \times 1}$.

While previous object-centric learning methods~\cite{zhou2023objects} directly apply $M$ to the input $V$ to remove redundant visual tokens, we find empirically that applying the mask to the visual embeddings instead produces better results. We hypothesize that although most of the visual background is redundant, there is still some useful semantic information (e.g. it may be useful to know if the action occurs in a kitchen, outdoors, etc.). Accordingly, the self-attention of the encoder instills some of that background information into each of the patch features. We apply the mask to $e_v^*$ by first patchifying $M$ to get $\tilde{M} \in\mathbb{R}^{T \times N \times 1}$. Then, we can directly apply the patchified mask elementwise $\tilde{e}_v = \tilde{M} * e_v^* \in\mathbb{R}^{T \times N \times D}$ so that only the patch features directly associated with the object regions are non-zero. We then apply mean-pooling on $\tilde{e}_v$ over the non-zero features across $T$ and $N$ to get a single embedding vector for the visual input $e_v \in \mathbb{R}^{D}$. The process for calculating $e_v$ is illustrated in \cref{fig:visual_framework}. 

\subsection{Training framework} \label{sec:training_framework}

Once we have our per-modality embeddings $e_v, e_a, e_l$, we first pretrain our model using the multimodal contrastive-consensus coding (MC3) loss introduced by Chen \etal~\cite{chen2024soundingactions} for their sounding action discovery task. It involves a two-stage training framework where first the embeddings are aligned contrastively (the ``align'' stage) before applying an additional consensus coding loss to refine the embeddings within each sample (the ``refine'' stage). After pretraining, we evaluate our model as is on sounding action discovery (\cref{sec:soundactions_experiment}). Then, we propose a finetuning method using a hard-negatives contrastive loss for sounding object detection (the ``finetune'' stage).


\paragraph{Align stage} 

The align stage produces an initial embedding space where different modality embeddings that capture the same interactions are closer together compared to embeddings from a different category of interaction. This is done by treating modalities belonging to the same data sample as a positive pair and contrasting that against negative pairs formed from the other samples in the batch. We use InfoNCE loss~\cite{oord2019representation} for this stage. Let $e_{x}^i$ be the embedding of modality $x$ for the $i^{th}$ sample in the batch $\mathcal{B}$. Then for a given modality pair $(x,y)$, the loss is:
\begin{equation}
  \mathcal{L}_{x\rightarrow y} = -\frac{1}{|\mathcal{B}|} \sum_{i \in \mathcal{B}} \log \frac{\exp(e_x^i e_y^i / \tau)}{\sum_{l \in \mathcal{B}} \exp(e_x^i e_y^l / \tau)}
  \label{eq:infonce_xy}
\end{equation}
and similarly, a symmetric version of \cref{eq:infonce_xy} is defined as:
\begin{equation}
  \mathcal{L}_{y\rightarrow x} = -\frac{1}{|\mathcal{B}|} \sum_{i \in \mathcal{B}} \log \frac{\exp(e_y^i e_x^i / \tau)}{\sum_{l \in \mathcal{B}} \exp(e_y^i e_x^l / \tau)}
  \label{eq:infonce_yx}
\end{equation}
where $\tau$ is the temperature. The final loss for this stage then is the sum of these two symmetric losses for every modality pair: $\mathcal{L_{\text{align}}} = \sum_{x,y} \mathcal{L}_{x\rightarrow y} + \mathcal{L}_{y\rightarrow x}$. Since we have three modalities, there are three pairs that we sum over.

\paragraph{Refine stage} 

The next stage introduces another loss called the multimodal consensus coding loss. At a high level, there may be scenarios where not every similarity score between the different modality pairs agree within the same video. This consensus loss penalizes samples whose scores do not agree and forces them to have a similar value. First, an anchor modality $A$ is chosen and a consensus score is calculated across each modality against $A$: 
\begin{equation}
    c^i = \mathcal{K}^{-1} \left( \min_{x, x \neq A} \left( \mathcal{K}_1(e_1^i e_A^i), \dots, \mathcal{K}_n(e_n^i e_A^t) \right) \right)
    \label{eq:consensus_score}
\end{equation}
where $\mathcal{K}_x(t) = \left( (t + 1) / 2 \right)^{\alpha_x}, t \in [-1, 1]$ and $\mathcal{K}^{-1}$ is the inverse function and $\alpha_x$ is a hyperparameter. We follow~\cite{chen2024soundingactions} and use audio as the anchor and set $\alpha_l=1$ and $\alpha_v=0.5$. The consensus score will be high only if all pairwise modality scores are high. If even one score does not agree, then the consensus score will be low. The corresponding loss forces all pairwise scores to follow this consensus:
\begin{equation}
    \mathcal{L}_{\text{consensus}} = \frac{1}{|\mathcal{B}|} \sum_{i \in \mathcal{B}} \sum_{x, x \neq a} \| e_x^i e_A^i - c^i \|_2
    \label{eq:consensus_loss}
\end{equation}
The final loss for the refine stage is then $\mathcal{L_{\text{refine}}} = \mathcal{L}_{\text{align}} + \mathcal{L}_{\text{consensus}}$.

\paragraph{Finetune stage}
After pretraining with the two-stage loss, the model can be directly evaluated on sounding action discovery. However, we notice that the MC3 loss alone is not conducive to localized understanding, which is important for sounding object detection. 
We hypothesize that although we introduce an object-aware approach, sounding action discovery is inherently a global task and there is nothing in the training yet that penalizes the model for attending to background regions. Accordingly, we introduce a hard-negatives contrastive loss to finetune our model for sounding object detection. We go back to the InfoNCE-based loss of the align stage, except we use additional audiovisual hard negatives. This is illustrated in \cref{fig:visual_framework}. Originally, negative pairs are taken from the $e_v$ embeddings of other samples in the batch. Now, we use background region embeddings from each sample as negative pairs as well. Formally one side of this symmetric loss is:
\begin{equation}
  \hat{\mathcal{L}}_{a\rightarrow v} = -\frac{1}{|\mathcal{B}|} \sum_{i \in \mathcal{B}} \log \frac{\exp(e_a^i e_v^i / \tau)}{\splitfrac{\sum_{l \in \mathcal{B}} \exp(e_a^i e_v^l / \tau)}{+ \sum_{m \in \mathcal{B}} \exp(e_a^i \hat{e}_v^m / \tau)}}
  \label{eq:finetune_xy}
\end{equation}
where $\hat{e}_v^i$ is the visual embedding from randomly sampling $\beta \%$ of the non-object regions of video clip $i$. For all models, we set $\beta$ to $50$. We also define $\hat{\mathcal{L}}_{v\rightarrow a}$ for symmetry and sum the two as our final finetuning loss: $ \mathcal{L}_{\text{finetune}} = \hat{\mathcal{L}}_{v\rightarrow a} + \hat{\mathcal{L}}_{a\rightarrow v}$.
Note that for the finetuning stage, we do not use the language modality since sounding object detection is a purely audiovisual task.

\subsection{Implementation details}

\paragraph{Encoders} For our visual encoder, we use a slot attention model pretrained on MS COCO 2017~\cite{lin2015microsoftcoco} for unsupervised object segmentation~\cite{Kakogeorgiou2024SPOT}. This architecture contains a bottleneck slot attention module that compresses input features into a small number of slot vectors before being decoded back out to the original features. Each slot learns to attend to regions that correspond to distinct objects (\cref{fig:object_detection}). We find that this provides a useful object prior from which we can train our model.
The slot attention encoder we use contains 7 slots. We find empirically that keeping the slot attention model's own encoder and slots frozen while only training its decoder leads to the best results. For the audio modality, following~\cite{chen2024soundingactions}, we use AST~\cite{gong21b_interspeech} which has been pretrained on ImageNet~\cite{russakovsky2015imagenet}. Finally, we use a frozen CLIP~\cite{radford2021learning} model to extract language features. 

\vspace{-2mm}
\paragraph{Training details} We train all models on four A40 GPUs with a per-GPU batch size of 32, leading to an effective batch size of 128. Following Chen \etal~\cite{chen2024soundingactions}, we train for five epochs each for the align and refine stages, using the final checkpoint for sounding action discovery. We then finetune for five epochs for sounding object detection. More details can be found in~\cref{app:implementation}.

%% file: sec/5_datasets.tex
\section{Training and evaluation data}
\label{sec:_datasets}

\begin{table}[t]
    \centering
    \begin{tabular}{cccccc}
        \toprule
         & & \multicolumn{2}{c}{\textit{\# Clips}} & \\
        \textbf{Task} & \textbf{Dataset} & \textbf{Train} & \textbf{Eval} \\
        \midrule
         \multirow{2}{*}{Object Det.} & Epic Kitchens & 67K & \colorbox{SkyBlue}{583} \\
         & Ego4D & 240K & \colorbox{SkyBlue}{572} \\
        
    \midrule
         Action Disc. & Ego4D & 240K & 20K \\
        \bottomrule
    \end{tabular}
    \caption{Summary of the train and evaluation splits of the datasets used for our two tasks. Splits in \colorbox{SkyBlue}{blue} indicate the object masks are manually annotated from our benchmark, while the remaining splits use our automatic process to calculate object masks.}
    \vspace{-3mm}
    \label{tab:dataset_summary}
\end{table}

\paragraph{Datasets}

Ego4D~\cite{grauman2022ego4d} is a dataset of egocentric videos over 3,600 hours long depicting everyday activities and collected around the world. A subset of the videos contain both audio and timestamped text narrations, consisting of sentences describing the actions of the camera-wearer.~\cite{chen2024action2sound} curated a cleaner subset of Ego4D with high action-audio correspondence called Ego4D-Sounds. We sample 240K clips from Ego4D-Sounds to use for pretraining, with each clip being 1.5s long. The clips also come with audio and text narrations. \cref{sec:automatic_labelling} describes how we automatically annotate this training set with object segmentation masks to use in our object-aware training framework.

For sounding object detection, we then finetune and evaluate our model on Ego4D and Epic Kitchens~\cite{Damen2021PAMI}. The latter features egocentric videos of kitchen activities paired with audio and timestamped text narrations. We finetune on the training split of each dataset and evaluate on our benchmark, where we manually annotate 2.5K clips for sounding actions and sounding objects, detailed in \cref{sec:object_detection_benchmark}. For sounding action discovery, we evaluate models after pretraining on Ego4D only using the sounding action annotations collected by~\cite{chen2024soundingactions}. We sample 1.6K clips for validation and 20K clips for test. We use our automatic object mask annotation procedure for this task. A summary of the datasets used for each task is shown in \cref{tab:dataset_summary}.




\subsection{Automatic object mask annotation} \label{sec:automatic_labelling}

Recall from \cref{sec:architecture} that we use object segmentation masks to guide our object-aware learning framework. Given the scale of data we train and finetune on, manually annotating ground truth masks is impractical. Instead, we leverage the fact that most interactions in egocentric videos involve hands acting on the objects. Using an off-the-shelf hand-object interaction (HOI) detection model~\cite{Shan2020understanding}, we label every fifth frame, extracting bounding boxes for hands and the objects they touch. We keep only the object bounding boxes and pass them to SAM 2~\cite{ravi2024sam2} which propagates these bounding boxes into segmentation masks for every video frame. Finally, we use non-maximum supression to filter out duplicate masks in each frame. We assume a maximum of two objects per interaction, so we sample the first two object masks if more than two remain after NMS. 

\subsection{Sounding object detection benchmark} \label{sec:object_detection_benchmark}

\begin{figure}[t]
  \centering
   \includegraphics[width=\columnwidth]{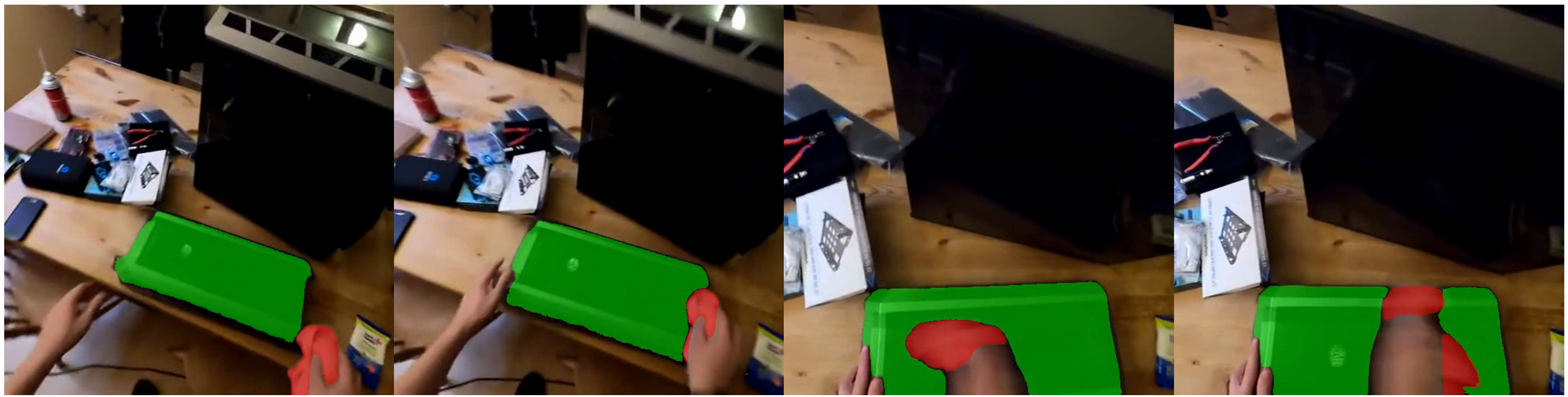}

   \caption{Examples of annotated frames from Ego4D from our benchmark, visualizing segmentation masks (red and green) of ground truth objects.}
   \label{fig:seg_mask_frames}
\end{figure}

In the previous section, we developed an automatic process for annotating object masks for our training and finetuning data (\cref{sec:automatic_labelling}) but it is based on the assumption that object interactions occur in the presence of hand contact. While this assumption works well for large-scale training data, we ideally want ground truth object masks when evaluating sounding object detection. Accordingly, we collect manual object segmentation mask annotations from the test splits of Epic Kitchens~\cite{Damen2021PAMI} and Ego4D~\cite{grauman2022ego4d}. 

First, annotators determine whether the action is sounding. If not, then no further steps are taken. If it is, they select the two objects that are directly involved by annotating keypoints across multiple frames and also label the name of each object. We then use SAM 2 to propagate the keypoints into segmentation masks for all frames in the video. There are certain object interactions where the sound mainly comes from a single object (e.g. opening a fridge door). Since we enforce the rule during annotation that two objects must be labelled, in these cases, we instruct the annotators to label the hands as the second object. But in the cases where there are two non-hand objects involved (e.g. knife on chopping board), the hand is not annotated. During evaluation however, we remove the hand masks, meaning the models do not need to detect hands as a sounding object since they are not included during training. More details about the data annotation process can be found in~\cref{app:annotation}. In total, we annotated 2.5K clips across the two datasets, of which 1.1K contain sounding actions paired with ground truth object masks. 

During evaluation, we create a pool of candidate object masks for each video by prompting OWLv2~\cite{minderer2023scaling} with the 521 nouns in the Ego4D taxonomy and filtering the bounding box detections based on a threshold of the confidence scores. We then use SAM 2 to get segmentation masks for all frames. To annotate the positive masks, we calculate the IoU between every pair of candidate and ground truth object mask and take the candidate mask with the highest IoU as the positive object. If none of the candidate masks match well (i.e. OWLv2 did not detect the sounding object), we simply add the ground truth mask to the pool.

%% file: sec/6_experiments.tex
\section{Experiments}
\label{sec:_experiments}

We evaluate our model and baselines and present ablation studies for two tasks: sounding object detection on Epic Kitchens and Ego4D and sounding action discovery on Ego4D. We show that in both cases, our model outperforms the previous state of the art methods.

\subsection{Sounding object detection}

\begin{figure*}[ht]
  \centering
   \includegraphics[width=0.95\linewidth]{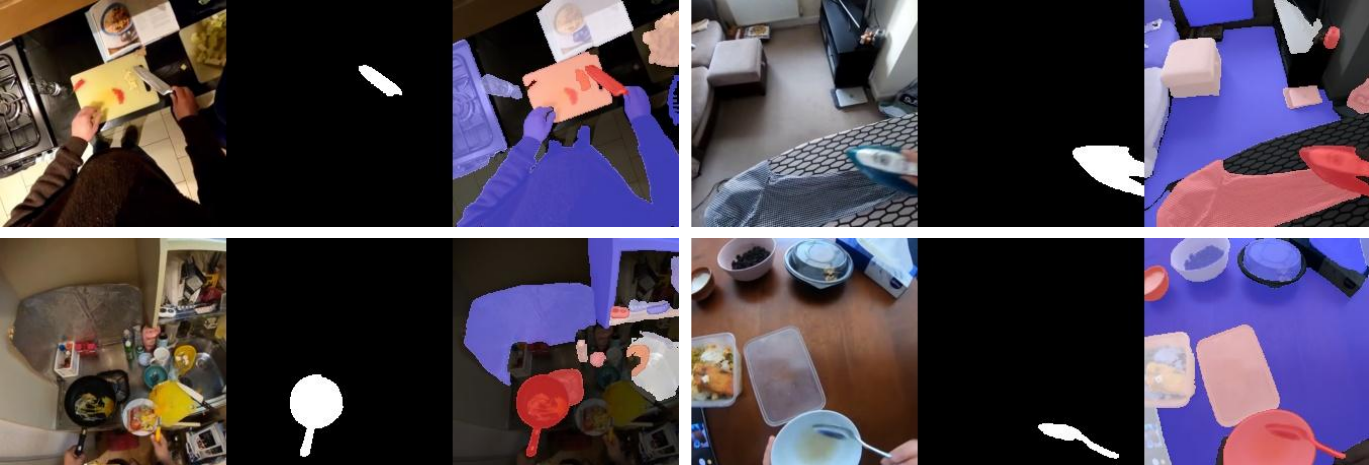}
   \caption{Qualitative results of our sounding object detection task. For each sub-figure from left to right: a) the original video frame, b) the ground truth object segmentation mask, and c) the audiovisual similarity score of each object region. The object regions in c) are detected using OWLv2~\cite{minderer2023scaling} and SAM 2~\cite{ravi2024sam2}. \textcolor{Blue}{\textbf{Dark blue}} is the lowest score and \textcolor{BrickRed}{\textbf{dark red}} is the highest. Refer to~\cref{app:colormap} for the colormap.}
   \label{fig:object_detection}
\end{figure*}

In this task, given the sound of an object interaction, a video frame depicting the action, and a pool of object masks, the model first computes a similarity score for each image patch by calculating the cosine similarity between the patch and audio embeddings. The similarity map is  interpolated to the original video dimensions and pixel-level similarity scores for each object gets mean-pooled. The object with the highest pooled score is taken as the model's prediction for the sounding object. If that object matches with one of the ground truth objects, then it is considered a success. We finetune our model on the train split of each dataset using our hard negatives contrastive loss after pretraining with the MC3 loss  (\cref{sec:training_framework}). Baselines contain both finetuned and zero-shot methods and we report the top-1 accuracy.

\vspace{-3mm}
\paragraph{Baselines} We compare our method with two categories of models. First, we compare against SoundingActions~\cite{chen2024soundingactions}, a model that has been pretrained on the same Ego4D data with the same MC3 loss. This baseline highlights the benefits of our object-aware approach compared to learning from global representations. Second, we compare with unsupervised audiovisual localization models. DenseAV~\cite{hamilton2024separating} is an audiovisual segmentation model that can localize both sounds and speech, SLAVC~\cite{mo2022SLAVC} uses a popular localization framework called multiple instance contrastive learning~\cite{mo2022localizing}, and SSLAlign~\cite{senocak2023sound} is a state of the art method. SoundingActions is finetuned using the hard negatives loss and SLAVC is finetuned using its original objective. We also include a naive vision-only baseline using the HOI detection model from~\cref{sec:automatic_labelling}. We match up to 2 object masks from the HOI model to the ground truth masks using IoU, counting an IoU $\geq$ 0.85 for at least one pair as a success. 

\paragraph{Results}

\begin{table}[t]
  \centering
  \setlength{\tabcolsep}{3pt}
  \renewcommand{\arraystretch}{1}
  \begin{tabular}{lccc}
    \toprule
         &  FT?  & Epic Kitchens & Ego4D \\
    \midrule
    Random & - & 17.7 & 17.1 \\
    Vision-only & - & 26.2 & 26.4\\
    \midrule
    DenseAV~\cite{hamilton2024separating} & \xmark & 26.4 & 29.7\\
    \midrule 
    SSLAlign~\cite{senocak2023sound} & \xmark & 39.5 & 33.9\\
    \midrule
    SLAVC~\cite{mo2022SLAVC} & \xmark & 34.8 & 33.0 \\
    SLAVC~\cite{mo2022SLAVC} & \cmark & 35.8 & 31.5 \\
    \midrule
    SoundingActions~\cite{chen2024soundingactions} & \cmark & 41.0 & 35.1 \\
    \midrule
    No object masking & \cmark & 46.3 & 40.4 \\
    No slot attention & \cmark & 46.1 & 43.0 \\
    Ours~\cite{yang2025clink} & \cmark & 49.6 & 43.9\\
    Ours++$^*$ & \cmark & \textbf{52.8} & \textbf{44.9}\\
    \bottomrule
  \end{tabular}
  \caption{Top-1 accuracy (\%) for sounding object detection. We finetune (FT) our model and the previous state of the art~\cite{chen2024soundingactions} from sounding action discovery pretraining. We also compare against finetuned and zero-shot audiovisual localization methods. $^*$\textit{Ours++} has the same model weights as \textit{Ours} but the slot attention visual encoder uses all sequence permutations~\cite{Kakogeorgiou2024SPOT} during eval rather than just the standard. Additional details in~\cref{app:permutation}.}
  \vspace{-3mm}
  \label{tab:object_detection}
\end{table}

\cref{tab:object_detection} shows the top-1 accuracy for sounding object detection across different methods and datasets. All three audiovisual localization baselines perform better than the naive baselines, demonstrating that their specialized frameworks can transfer over to understanding correspondences between object interactions and sound. Out of the three methods, DenseAV is the weakest model, likely due to its ability to localize for both sound and speech. This can add an additional layer of complexity since the model must also predict which class the audio input belongs to. And for all three localization methods, their training data~\cite{aytar2016soundnet, chen2020vggsound, gemmeke2017audioset} falls under predefined audio classes (e.g. animals, music, etc.). On the other hand, the sounds of object interactions can be more subtle and exist on the long tail of data seen by these methods. For most methods, Ego4D is a harder dataset than Epic Kitchens, since the former consists of a wide variety of scenarios and objects while the latter is limited to the kitchen setting.

Next, although SoundingActions uses the same training loss and data as our method, we achieve significantly better performance on this task by 11.8\% on Epic Kitchens and 9.8\% on Ego4D. Unlike the baselines, our method is object-aware, incorporating object masks that require no supervision and leveraging object priors from a pretrained slot attention encoder. This guides our model to focus on the objects relevant to the interaction. Consequently, our model better captures the subtle correlations between the sounds of different object interactions. Finally, we show qualitative examples from our model in \cref{fig:object_detection}. We visualize the relative rankings of the object scores, with \textcolor{Blue}{\textbf{dark blue}}
being the least correlated with the audio and \textcolor{BrickRed}{\textbf{dark red}} the most. Additional results can be found in~\cref{app:sounding_object_qualitative}.

\vspace{-1mm}
\paragraph{Ablations}
We present two ablations: 1) with no object-aware masking but still using the slot attention encoder and 2) without slot attention but still using object-aware masking. The latter is equivalent to training SoundingActions with the addition of object-aware masking. Results are shown in \cref{tab:object_detection}. We see that applying either method individually already leads to improvements over the baselines, validating their effectiveness. However, using them jointly, as our final model does, ultimately leads to the best results.


\subsection{Sounding action discovery} \label{sec:soundactions_experiment}

\begin{table}[t]
    \centering
    \begin{tabular}{lcccccc}
        \toprule
        \textbf{} & \multicolumn{2}{c}{\textit{AV}} & \multicolumn{2}{c}{\textit{AL}} \\
        & ROC & PR & ROC & PR \\
        \midrule
        Random & 0.500 & 0.587 & 0.500 & 0.587 \\
        ImageBind~\cite{girdhar2023imagebind} & 0.524 & 0.611 & 0.527 & 0.632 \\
        LanguageBind~\cite{zhu2023languagebind} & 0.529 & 0.627 & 0.551 & 0.636 \\
        SoundingActions~\cite{chen2024soundingactions} & 0.592 & 0.683 & 0.616 & 0.720 \\
        \midrule
        No object masking & 0.593 & 0.696 & 0.618 & 0.720 \\
        No slot attention & 0.578 & 0.671 & 0.607 & 0.713 \\
        Ours & \textbf{0.617} & \textbf{0.706} & \textbf{0.630} & \textbf{0.726} \\
        \bottomrule
    \end{tabular}
    \caption{Sounding action discovery results for audio-vision (AV) and audio-language (AL). We report area-under-curve values for receiver operating characteristic (ROC) and precision-recall (PR) curves. We compare against the previous state of the art~\cite{chen2024soundingactions} along with zero-shot contrastive methods~\cite{girdhar2023imagebind, zhu2023languagebind}. We empirically found that the changes made in \textit{Ours++} made no difference for this task.}
    \label{tab:action_discovery}
\end{table}

Next, we evaluate our model on sounding action discovery. After pretraining our model on Ego4D with the MC3 loss, we compute vision, audio, and language embeddings for each of the 20K test clips and calculate pairwise cosine similarity scores for audio-vision (AV) and audio-language (AL). We follow Chen \etal~\cite{chen2024soundingactions} and report area under the receiver operating characteristic curve (AUC-ROC) and area under the precision-recall curve (AUC-PR).

\paragraph{Baselines}

We compare our method against SoundingActions, which was the previous state of the art on this task. We also compare against two other multimodal foundational models: ImageBind~\cite{girdhar2023imagebind} and LanguageBind~\cite{zhu2023languagebind}. Both models are trained contrastively on Internet-scale data to align a wide range of modalities, including vision, audio, language, depth, and thermal, into a single embedding space. Given the scale of data these models have seen, we evaluate these methods zero-shot.

\paragraph{Results}

\cref{tab:action_discovery} shows AUC-ROC and AUC-PR results for sounding action discovery for both audio-vision (AV) and audio-language (AL) modality pairs. We see that both ImageBind and LanguageBind perform marginally better than chance, indicating the large-scale pretraining of these models instills some knowledge of sounding actions. SoundingActions, with its MC3 loss, does better than the other baselines because the loss supervises the consensus between all three modalities to be better aligned. While our model uses the same loss, we further improve its performance by integrating object-aware training. This demonstrates that even for a more global reasoning task, a localized approach that prioritizes relevant objects is essential. 

Out of the two modality pairs, AV benefits the most from the performance boost provided by our object-aware framework. Since our approach focuses on the visual modality, it is reasonable that AL sounding action discovery does not benefit equally. The improved performance nonetheless suggests that having a more compact and expressive visual feature helps the model learn better representations for the other modalities as well, since the contrastive loss iterates through all modality pairs.

\begin{figure}[t]
  \centering
   \includegraphics[width=\columnwidth]{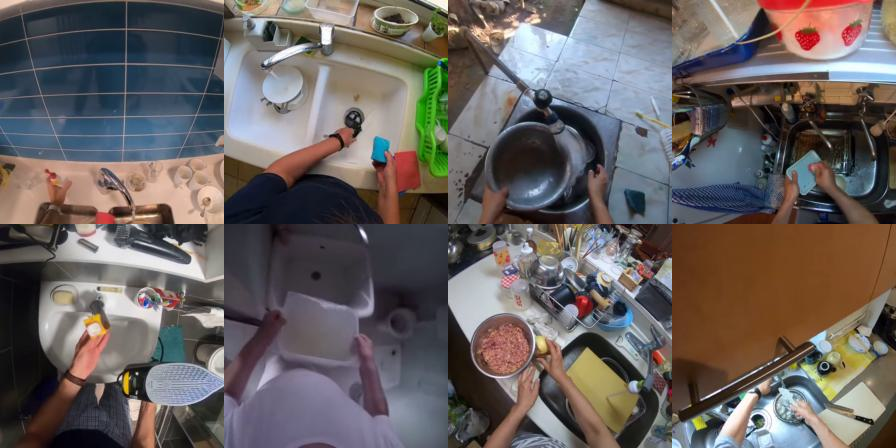}

   \caption{Frames from the same cluster of visual embeddings output by our model from Ego4D. The frames cover different background appearances but all correspond with the sound of flowing water, either from a tap or being poured out from a container.}
   \label{fig:clustering}
\end{figure}

To further demonstrate the audiovisual correspondences learned by our model, we follow~\cite{chen2024soundingactions} and perform agglomerative clustering on our visual embeddings using 20 clusters. \cref{fig:clustering} shows eight examples from one such cluster. We see that the images cover a wide variety of perspectives and visual backgrounds with the underlying similarity being that these videos all contain the sound of flowing water. This includes water running from the tap but also water being poured out of a container (second example in the bottom row). We include more cluster examples in~\cref{app:clustering}.

\paragraph{Ablations}

We perform the same ablation studies as sounding object detection and report results in \cref{tab:action_discovery}. Without object masking, we get minimal improvements in performance compared to SoundingActions while applying object masking directly to SoundingActions actually leads to worse results.

%% file: sec/7_conclusion.tex
\section{Conclusion}
\label{sec:_conclusion}

We investigate how to use egocentric videos of daily actions to learn the relationship between object interactions and their sounds. We train a multimodal object-aware model capable of detecting the \textit{sounding object} from a set of candidates present in a scene and distinguishing whether an action is sounding or not. Our framework focuses on the important regions of a video, guided during training by object segmentation masks computed using an automatic pipeline, allowing us to learn a more compact and expressive representation that outperforms previous multimodal methods.


%% file: sec/8_supplementary.tex
\appendix



    



\section{Additional sounding object detection qualitative results}\label{app:sounding_object_qualitative}

\begin{figure*}[t]
  \centering
   \includegraphics[width=\linewidth]{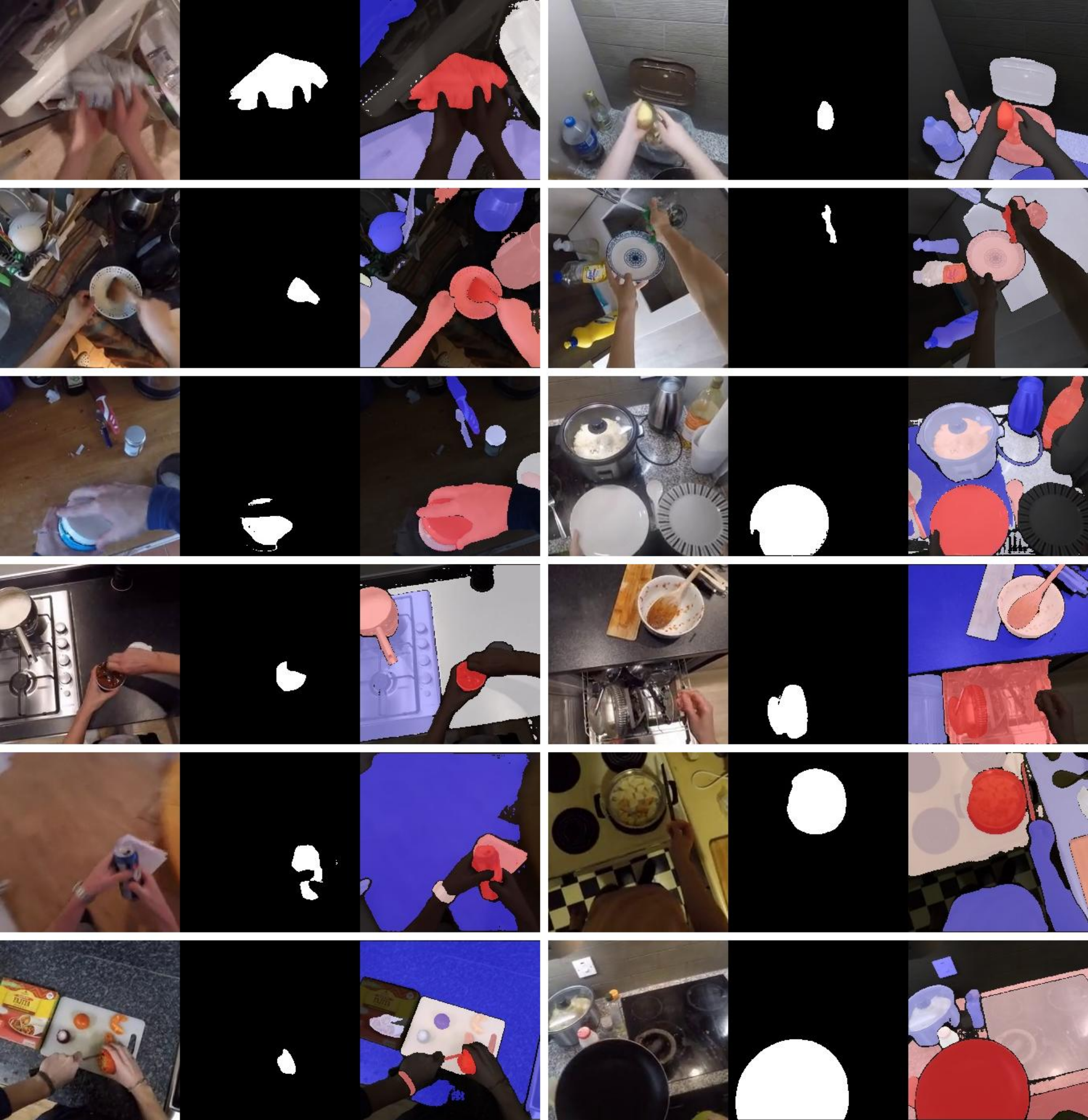}
   \caption{Additional qualitative results for sounding object detection on Epic Kitchens.}
   \label{fig:epic_kit_object_qual}
\end{figure*}

\begin{figure*}[t]
  \centering
   \includegraphics[width=\linewidth]{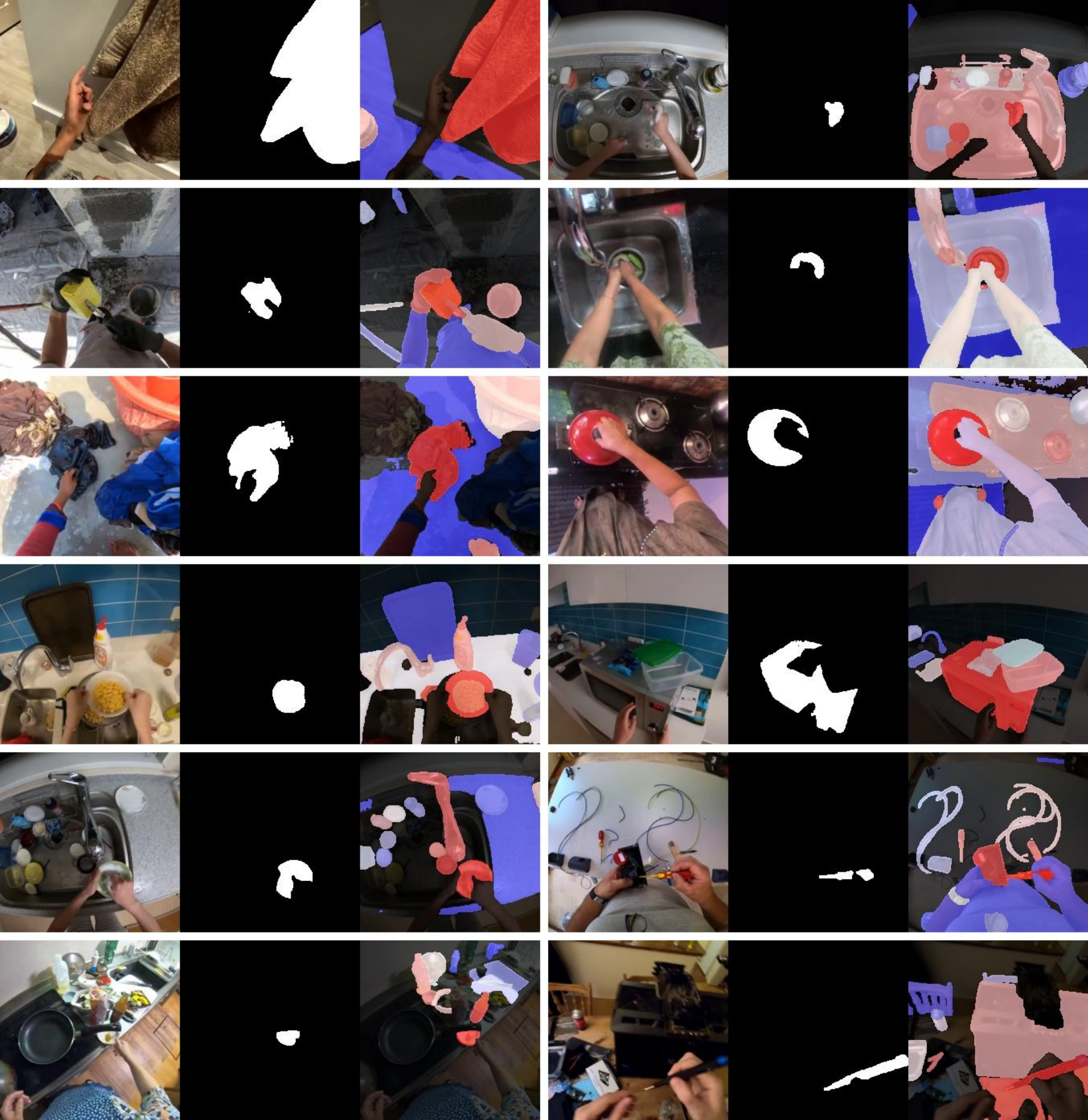}
   \caption{Additional qualitative results for sounding object detection on Ego4D.}
   \label{fig:ego4d_object_qual}
\end{figure*}

We provide additional qualitative results for sounding object detection on Epic Kitchens in \cref{fig:epic_kit_object_qual} and Ego4D in \cref{fig:ego4d_object_qual}.

\section{Additional visual clustering results}\label{app:clustering}

\begin{figure*}[t]
  \centering
  \begin{subfigure}{1.0\columnwidth}
    \includegraphics[width=1.0\columnwidth]{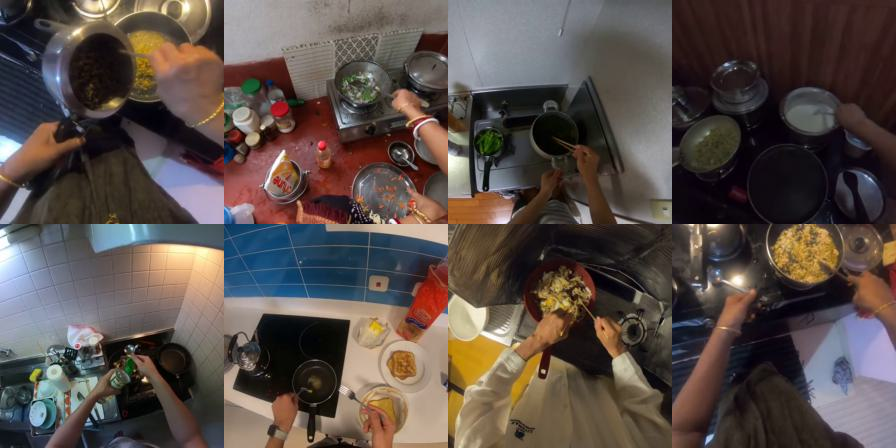}
    \caption{Visual embeddings that correspond to sounds of food sizzling.}
    \label{fig:clustering_sizzling}
  \end{subfigure}
  \hfill
  \begin{subfigure}{1.0\columnwidth}
    \includegraphics[width=1.0\columnwidth]{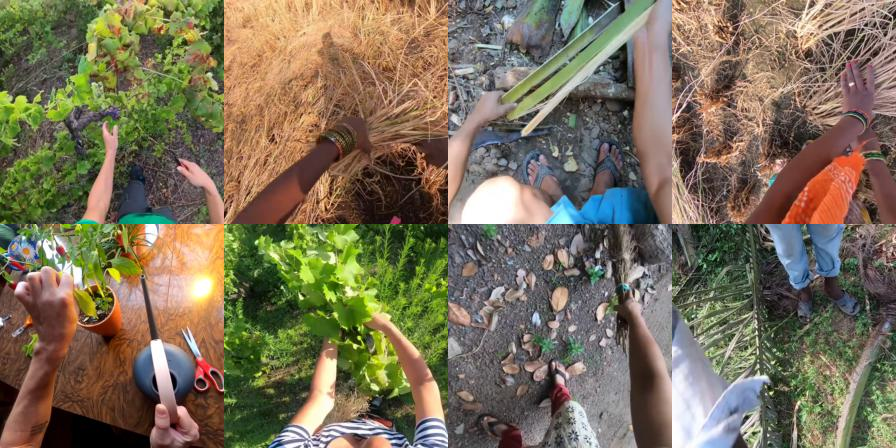}
    \caption{Visual embeddings that correspond to sounds of plants rustling.}
    \label{fig:clustering_rustling}
  \end{subfigure}
  \caption{Additional visual embedding clustering results. Each cluster shown corresponds to visual frames with diverse perspectives and backgrounds. But the common trait is all corresponding sounds belong to the same category.}
  \label{fig:clustering_supp}
\end{figure*}

Additional examples of visual embedding clusters from Ego4D are shown in \cref{fig:clustering_supp}. In each case, cluster examples correspond to scenarios with similar sounds.

\section{Similarity map color legend}\label{app:colormap}

\begin{figure}
  \centering
   \includegraphics[width=0.1\linewidth]{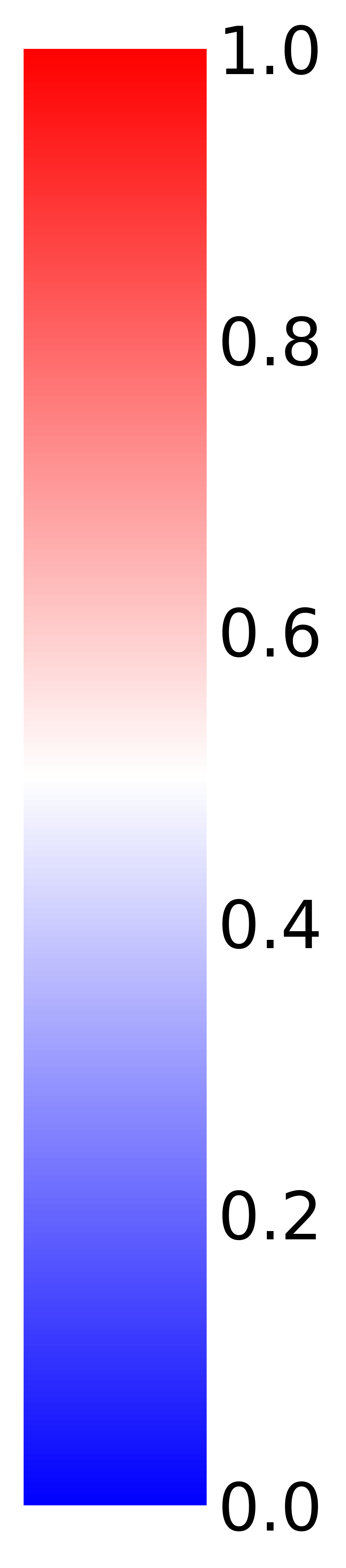}
   \caption{Colorbar legend used to visualize object region scores in sounding object detection.}
   \label{fig:colorbar}
\end{figure}

\cref{fig:colorbar} shows the color scale used for visualizing the similarity maps for sounding object detection. Darker colors correspond to the extremes with blue being a low value and red being a high value. Since similarity is calculated using cosine similarity between the vision and audio embeddings, the scores are in the range $[0,1]$ with a higher value depicting greater correspondence.

\section{Annotation details for the sounding object detection benchmark}\label{app:annotation}

\begin{figure*}
  \centering
   \includegraphics[width=0.9\linewidth]{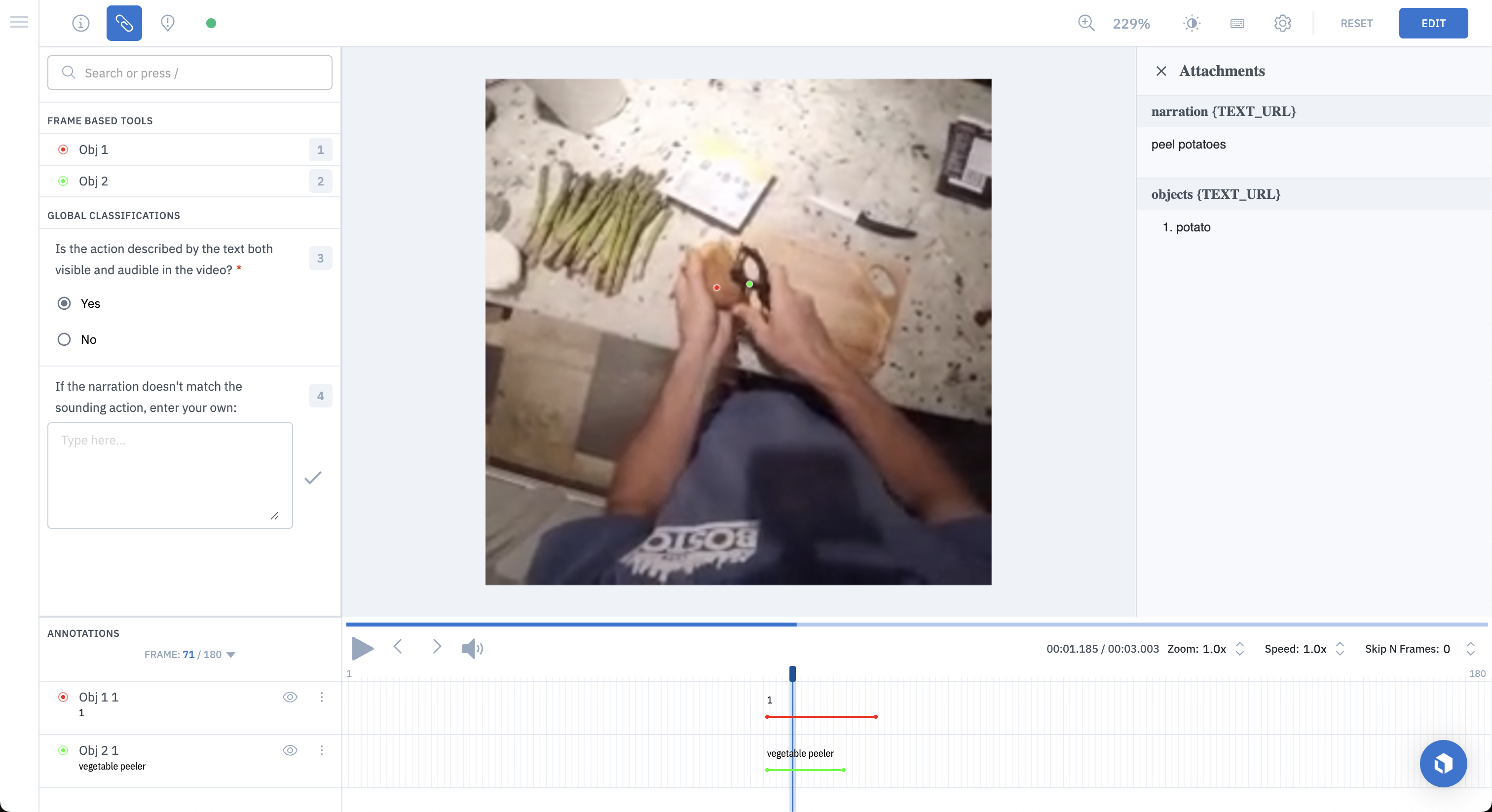}
   \caption{Screenshot of the Labelbox~\cite{labelbox2025} interface used to annotate ground truth object masks for our sounding object detection benchmark. In addition to answering the questions in the left column, annotators can scrub through individual frames and apply keypoints to the objects involved in the action. These keypoints are then used with SAM 2~\cite{ravi2024sam2} to extract ground truth object masks.}
   \label{fig:labelbox}
\end{figure*}

We use Labelbox~\cite{labelbox2025} to develop our annotation interface, shown in \cref{fig:labelbox}. First, annotators answer whether the action is sounding or not. If the action is not sounding, then no further labels are required. If the action is sounding, they then label the locations of the two objects involved by placing keypoints to track the objects across multiple frames. The objects are also labelled with either a noun present in the provided narration or a text input field is provided for the annotators to describe the objects and optionally provide a more descriptive narration.

\section{Additional implementation details}\label{app:implementation}

We use a learning rate of 5e-5 and 4 video frames per clip during pretraining. Each frame is resized to 224 on the smaller edge and then center-cropped to 224$\times$224. We use a patch size of 16$\times$16. During training, the 4 frames are sampled randomly. During sounding action discovery evaluation, the 4 frames are sampled uniformly. Meanwhile, sounding object detection uses 1 frame sampled from the middle of the clip. 

For the audio encoder, we use AST~\cite{gong21b_interspeech} pretrained on ImageNet~\cite{russakovsky2015imagenet}. The input to the audio encoder are fbank features calculated on the waveform using 128 Mel frequency bins, 10ms frame shift, and a Hanning window. We use a sample rate of 16kHz. 

For the language encoder, we use the pretrained CLIP model from Huggingface, specifically ``openai/clip-vit-base-patch32'', which we keep frozen. 

For our visual encoder, we initialize from a pretrained slot attention model trained on MS COCO 2017~\cite{lin2015microsoftcoco} from~\cite{Kakogeorgiou2024SPOT} that uses 7 slots. We keep the encoder and slot embeddings of the slot attention encoder frozen and only train the decoder weights. 

We project all modalities into a common 256-dimensional embedding space. We use video clips that are 1.5s long, which was found to be the ideal length in~\cite{chen2024soundingactions}. Given the timestamp of the narration, we extract 0.5s from before and 1s after.

Finally, we use a confidence score threshold of 0.35 for OWLv2~\cite{minderer2023scaling} when detecting object candidates in a scene for sounding object detection.

\section{Sequence permutations for the slot attention encoder}\label{app:permutation}

One of the contributions from SPOT~\cite{Kakogeorgiou2024SPOT}, the slot attention model that we use as our visual encoder, is the introduction of a patch-order permutation strategy that changes the output sequence order of the autoregressive decoder. The authors found that the initial tokens rely heavily on information from slot vectors but as more tokens are decoded, the reliance on context from slot vectors diminish and thus provide weak supervisory signals for optimizing the slot vectors.

Changing the order, or permutation, of how tokens are autoregressively decoded during training introduces variability that teaches the model to rely more on slot information for all tokens, leading to better downstream performance. During test time, the model can be set to either use the standard permutation, a random permutation, or the set of all permutations used during training before averaging the outputs. The paper's codebase\footnote{https://github.com/gkakogeorgiou/spot} uses only the standard permutation during test time by default, but we empirically found that using all permutations leads to better results.

%% file: main.bbl
\begin{thebibliography}{48}
\providecommand{\natexlab}[1]{#1}
\providecommand{\url}[1]{\texttt{#1}}
\expandafter\ifx\csname urlstyle\endcsname\relax
  \providecommand{\doi}[1]{doi: #1}\else
  \providecommand{\doi}{doi: \begingroup \urlstyle{rm}\Url}\fi

\bibitem[Afouras et~al.(2020)Afouras, Owens, Chung, and Zisserman]{afouras2020self}
Triantafyllos Afouras, Andrew Owens, Joon~Son Chung, and Andrew Zisserman.
\newblock Self-supervised learning of audio-visual objects from video.
\newblock In \emph{Computer Vision--ECCV 2020: 16th European Conference, Glasgow, UK, August 23--28, 2020, Proceedings, Part XVIII 16}, pages 208--224. Springer, 2020.

\bibitem[Arandjelovic and Zisserman(2017)]{arandjelovic2017look}
Relja Arandjelovic and Andrew Zisserman.
\newblock Look, listen and learn.
\newblock In \emph{Proceedings of the IEEE international conference on computer vision}, pages 609--617, 2017.

\bibitem[Arnab et~al.(2021)Arnab, Dehghani, Heigold, Sun, Lučić, and Schmid]{arnab2021vivit}
Anurag Arnab, Mostafa Dehghani, Georg Heigold, Chen Sun, Mario Lučić, and Cordelia Schmid.
\newblock Vivit: A video vision transformer, 2021.

\bibitem[Aytar et~al.(2016)Aytar, Vondrick, and Torralba]{aytar2016soundnet}
Yusuf Aytar, Carl Vondrick, and Antonio Torralba.
\newblock Soundnet: Learning sound representations from unlabeled video.
\newblock \emph{Advances in neural information processing systems}, 29, 2016.

\bibitem[Bertasius et~al.(2021)Bertasius, Wang, and Torresani]{bertasius2021spacetime}
Gedas Bertasius, Heng Wang, and Lorenzo Torresani.
\newblock Is space-time attention all you need for video understanding?, 2021.

\bibitem[Chen et~al.(2024{\natexlab{a}})Chen, Ashutosh, Girdhar, Harwath, and Grauman]{chen2024soundingactions}
Changan Chen, Kumar Ashutosh, Rohit Girdhar, David Harwath, and Kristen Grauman.
\newblock Soundingactions: Learning how actions sound from narrated egocentric videos.
\newblock In \emph{CVPR}, 2024{\natexlab{a}}.

\bibitem[Chen et~al.(2024{\natexlab{b}})Chen, Peng, Baid, Xue, Hsu, Harwath, and Grauman]{chen2024action2sound}
Changan Chen, Puyuan Peng, Ami Baid, Zihui Xue, Wei-Ning Hsu, David Harwath, and Kristen Grauman.
\newblock Action2sound: Ambient-aware generation of action sounds from egocentric videos.
\newblock In \emph{European Conference on Computer Vision}, pages 277--295. Springer, 2024{\natexlab{b}}.

\bibitem[Chen et~al.(2020)Chen, Xie, Vedaldi, and Zisserman]{chen2020vggsound}
Honglie Chen, Weidi Xie, Andrea Vedaldi, and Andrew Zisserman.
\newblock Vggsound: A large-scale audio-visual dataset, 2020.

\bibitem[Chen et~al.(2021)Chen, Xie, Afouras, Nagrani, Vedaldi, and Zisserman]{chen2021localizing}
Honglie Chen, Weidi Xie, Triantafyllos Afouras, Arsha Nagrani, Andrea Vedaldi, and Andrew Zisserman.
\newblock Localizing visual sounds the hard way, 2021.

\bibitem[Damen et~al.(2021)Damen, Doughty, Farinella, Fidler, Furnari, Kazakos, Moltisanti, Munro, Perrett, Price, and Wray]{Damen2021PAMI}
Dima Damen, Hazel Doughty, Giovanni~Maria Farinella, Sanja Fidler, Antonino Furnari, Evangelos Kazakos, Davide Moltisanti, Jonathan Munro, Toby Perrett, Will Price, and Michael Wray.
\newblock The epic-kitchens dataset: Collection, challenges and baselines.
\newblock \emph{IEEE Transactions on Pattern Analysis and Machine Intelligence (TPAMI)}, 43\penalty0 (11):\penalty0 4125--4141, 2021.

\bibitem[Elsayed et~al.(2022)Elsayed, Mahendran, van Steenkiste, Greff, Mozer, and Kipf]{elsayed2022savi}
Gamaleldin~F. Elsayed, Aravindh Mahendran, Sjoerd van Steenkiste, Klaus Greff, Michael~C. Mozer, and Thomas Kipf.
\newblock Savi++: Towards end-to-end object-centric learning from real-world videos, 2022.

\bibitem[Gan et~al.(2019)Gan, Zhao, Chen, Cox, and Torralba]{gan2019self}
Chuang Gan, Hang Zhao, Peihao Chen, David Cox, and Antonio Torralba.
\newblock Self-supervised moving vehicle tracking with stereo sound.
\newblock In \emph{Proceedings of the IEEE/CVF international conference on computer vision}, pages 7053--7062, 2019.

\bibitem[Gao et~al.(2021)Gao, Chang, Mall, Fei-Fei, and Wu]{gao2021ObjectFolder}
Ruohan Gao, Yen-Yu Chang, Shivani Mall, Li Fei-Fei, and Jiajun Wu.
\newblock Objectfolder: A dataset of objects with implicit visual, auditory, and tactile representations.
\newblock In \emph{CoRL}, 2021.

\bibitem[Gao et~al.(2022)Gao, Si, Chang, Clarke, Bohg, Fei-Fei, Yuan, and Wu]{gao2022ObjectFolderV2}
Ruohan Gao, Zilin Si, Yen-Yu Chang, Samuel Clarke, Jeannette Bohg, Li Fei-Fei, Wenzhen Yuan, and Jiajun Wu.
\newblock Objectfolder 2.0: A multisensory object dataset for sim2real transfer.
\newblock In \emph{CVPR}, 2022.

\bibitem[Gemmeke et~al.(2017)Gemmeke, Ellis, Freedman, Jansen, Lawrence, Moore, Plakal, and Ritter]{gemmeke2017audioset}
Jort~F. Gemmeke, Daniel P.~W. Ellis, Dylan Freedman, Aren Jansen, Wade Lawrence, R.~Channing Moore, Manoj Plakal, and Marvin Ritter.
\newblock Audio set: An ontology and human-labeled dataset for audio events.
\newblock In \emph{Proc. IEEE ICASSP 2017}, 2017.

\bibitem[Girdhar et~al.(2023)Girdhar, El-Nouby, Liu, Singh, Alwala, Joulin, and Misra]{girdhar2023imagebind}
Rohit Girdhar, Alaaeldin El-Nouby, Zhuang Liu, Mannat Singh, Kalyan~Vasudev Alwala, Armand Joulin, and Ishan Misra.
\newblock Imagebind: One embedding space to bind them all, 2023.

\bibitem[Gong et~al.(2021)Gong, Chung, and Glass]{gong21b_interspeech}
Yuan Gong, Yu-An Chung, and James Glass.
\newblock {AST: Audio Spectrogram Transformer}.
\newblock In \emph{Proc. Interspeech 2021}, pages 571--575, 2021.

\bibitem[Grauman et~al.(2022)Grauman, Westbury, Byrne, Chavis, Furnari, Girdhar, Hamburger, Jiang, Liu, Liu, Martin, Nagarajan, Radosavovic, Ramakrishnan, Ryan, Sharma, Wray, Xu, Xu, Zhao, Bansal, Batra, Cartillier, Crane, Do, Doulaty, Erapalli, Feichtenhofer, Fragomeni, Fu, Gebreselasie, Gonzalez, Hillis, Huang, Huang, Jia, Khoo, Kolar, Kottur, Kumar, Landini, Li, Li, Li, Mangalam, Modhugu, Munro, Murrell, Nishiyasu, Price, Puentes, Ramazanova, Sari, Somasundaram, Southerland, Sugano, Tao, Vo, Wang, Wu, Yagi, Zhao, Zhu, Arbelaez, Crandall, Damen, Farinella, Fuegen, Ghanem, Ithapu, Jawahar, Joo, Kitani, Li, Newcombe, Oliva, Park, Rehg, Sato, Shi, Shou, Torralba, Torresani, Yan, and Malik]{grauman2022ego4d}
Kristen Grauman, Andrew Westbury, Eugene Byrne, Zachary Chavis, Antonino Furnari, Rohit Girdhar, Jackson Hamburger, Hao Jiang, Miao Liu, Xingyu Liu, Miguel Martin, Tushar Nagarajan, Ilija Radosavovic, Santhosh~Kumar Ramakrishnan, Fiona Ryan, Jayant Sharma, Michael Wray, Mengmeng Xu, Eric~Zhongcong Xu, Chen Zhao, Siddhant Bansal, Dhruv Batra, Vincent Cartillier, Sean Crane, Tien Do, Morrie Doulaty, Akshay Erapalli, Christoph Feichtenhofer, Adriano Fragomeni, Qichen Fu, Abrham Gebreselasie, Cristina Gonzalez, James Hillis, Xuhua Huang, Yifei Huang, Wenqi Jia, Weslie Khoo, Jachym Kolar, Satwik Kottur, Anurag Kumar, Federico Landini, Chao Li, Yanghao Li, Zhenqiang Li, Karttikeya Mangalam, Raghava Modhugu, Jonathan Munro, Tullie Murrell, Takumi Nishiyasu, Will Price, Paola~Ruiz Puentes, Merey Ramazanova, Leda Sari, Kiran Somasundaram, Audrey Southerland, Yusuke Sugano, Ruijie Tao, Minh Vo, Yuchen Wang, Xindi Wu, Takuma Yagi, Ziwei Zhao, Yunyi Zhu, Pablo Arbelaez, David Crandall, Dima Damen, Giovanni~Maria
  Farinella, Christian Fuegen, Bernard Ghanem, Vamsi~Krishna Ithapu, C.~V. Jawahar, Hanbyul Joo, Kris Kitani, Haizhou Li, Richard Newcombe, Aude Oliva, Hyun~Soo Park, James~M. Rehg, Yoichi Sato, Jianbo Shi, Mike~Zheng Shou, Antonio Torralba, Lorenzo Torresani, Mingfei Yan, and Jitendra Malik.
\newblock Ego4d: Around the world in 3,000 hours of egocentric video, 2022.

\bibitem[Grill-Spector and Kanwisher(2005)]{Grill-Spector2005-or}
Kalanit Grill-Spector and Nancy Kanwisher.
\newblock Visual recognition: as soon as you know it is there, you know what it is.
\newblock \emph{Psychol. Sci.}, 16\penalty0 (2):\penalty0 152--160, 2005.

\bibitem[Hamilton et~al.(2024)Hamilton, Zisserman, Hershey, and Freeman]{hamilton2024separating}
Mark Hamilton, Andrew Zisserman, John~R. Hershey, and William~T. Freeman.
\newblock Separating the "chirp" from the "chat": Self-supervised visual grounding of sound and language, 2024.

\bibitem[Kakogeorgiou et~al.(2024)Kakogeorgiou, Gidaris, Karantzalos, and Komodakis]{Kakogeorgiou2024SPOT}
Ioannis Kakogeorgiou, Spyros Gidaris, Konstantinos Karantzalos, and Nikos Komodakis.
\newblock Spot: Self-training with patch-order permutation for object-centric learning with autoregressive transformers.
\newblock In \emph{Proceedings of the IEEE/CVF Conference on Computer Vision and Pattern Recognition (CVPR)}, pages 22776--22786, 2024.

\bibitem[Korbar et~al.(2018)Korbar, Tran, and Torresani]{korbar2018cooperative}
Bruno Korbar, Du Tran, and Lorenzo Torresani.
\newblock Cooperative learning of audio and video models from self-supervised synchronization.
\newblock \emph{Advances in Neural Information Processing Systems}, 31, 2018.

\bibitem[Labelbox(2025)]{labelbox2025}
Labelbox.
\newblock Labelbox.
\newblock \url{https://labelbox.com}, 2025.
\newblock Online.

\bibitem[Lin et~al.(2015)Lin, Maire, Belongie, Bourdev, Girshick, Hays, Perona, Ramanan, Zitnick, and Dollár]{lin2015microsoftcoco}
Tsung-Yi Lin, Michael Maire, Serge Belongie, Lubomir Bourdev, Ross Girshick, James Hays, Pietro Perona, Deva Ramanan, C.~Lawrence Zitnick, and Piotr Dollár.
\newblock Microsoft coco: Common objects in context, 2015.

\bibitem[Liu et~al.(2021)Liu, Ning, Cao, Wei, Zhang, Lin, and Hu]{liu2021videoswin}
Ze Liu, Jia Ning, Yue Cao, Yixuan Wei, Zheng Zhang, Stephen Lin, and Han Hu.
\newblock Video swin transformer, 2021.

\bibitem[Locatello et~al.(2020)Locatello, Weissenborn, Unterthiner, Mahendran, Heigold, Uszkoreit, Dosovitskiy, and Kipf]{locatello2020object}
Francesco Locatello, Dirk Weissenborn, Thomas Unterthiner, Aravindh Mahendran, Georg Heigold, Jakob Uszkoreit, Alexey Dosovitskiy, and Thomas Kipf.
\newblock Object-centric learning with slot attention, 2020.

\bibitem[Minderer et~al.(2023)Minderer, Gritsenko, and Houlsby]{minderer2023scaling}
Matthias Minderer, Alexey Gritsenko, and Neil Houlsby.
\newblock Scaling open-vocabulary object detection.
\newblock In \emph{Advances in Neural Information Processing Systems}, pages 72983--73007. Curran Associates, Inc., 2023.

\bibitem[Mo and Morgado(2022{\natexlab{a}})]{mo2022SLAVC}
Shentong Mo and Pedro Morgado.
\newblock A closer look at weakly-supervised audio-visual source localization.
\newblock In \emph{Advances in Neural Information Processing Systems}, 2022{\natexlab{a}}.

\bibitem[Mo and Morgado(2022{\natexlab{b}})]{mo2022localizing}
Shentong Mo and Pedro Morgado.
\newblock Localizing visual sounds the easy way, 2022{\natexlab{b}}.

\bibitem[Radford et~al.(2021)Radford, Kim, Hallacy, Ramesh, Goh, Agarwal, Sastry, Askell, Mishkin, Clark, Krueger, and Sutskever]{radford2021learning}
Alec Radford, Jong~Wook Kim, Chris Hallacy, Aditya Ramesh, Gabriel Goh, Sandhini Agarwal, Girish Sastry, Amanda Askell, Pamela Mishkin, Jack Clark, Gretchen Krueger, and Ilya Sutskever.
\newblock Learning transferable visual models from natural language supervision, 2021.

\bibitem[Ravi et~al.(2024)Ravi, Gabeur, Hu, Hu, Ryali, Ma, Khedr, Rädle, Rolland, Gustafson, Mintun, Pan, Alwala, Carion, Wu, Girshick, Dollár, and Feichtenhofer]{ravi2024sam2}
Nikhila Ravi, Valentin Gabeur, Yuan-Ting Hu, Ronghang Hu, Chaitanya Ryali, Tengyu Ma, Haitham Khedr, Roman Rädle, Chloe Rolland, Laura Gustafson, Eric Mintun, Junting Pan, Kalyan~Vasudev Alwala, Nicolas Carion, Chao-Yuan Wu, Ross Girshick, Piotr Dollár, and Christoph Feichtenhofer.
\newblock Sam 2: Segment anything in images and videos, 2024.

\bibitem[Russakovsky et~al.(2015)Russakovsky, Deng, Su, Krause, Satheesh, Ma, Huang, Karpathy, Khosla, Bernstein, Berg, and Fei-Fei]{russakovsky2015imagenet}
Olga Russakovsky, Jia Deng, Hao Su, Jonathan Krause, Sanjeev Satheesh, Sean Ma, Zhiheng Huang, Andrej Karpathy, Aditya Khosla, Michael Bernstein, Alexander~C. Berg, and Li Fei-Fei.
\newblock Imagenet large scale visual recognition challenge, 2015.

\bibitem[Senocak et~al.(2018)Senocak, Oh, Kim, Yang, and Kweon]{senocak2018learning}
Arda Senocak, Tae-Hyun Oh, Junsik Kim, Ming-Hsuan Yang, and In~So Kweon.
\newblock Learning to localize sound source in visual scenes, 2018.

\bibitem[Senocak et~al.(2023)Senocak, Ryu, Kim, Oh, Pfister, and Chung]{senocak2023sound}
Arda Senocak, Hyeonggon Ryu, Junsik Kim, Tae-Hyun Oh, Hanspeter Pfister, and Joon~Son Chung.
\newblock Sound source localization is all about cross-modal alignment.
\newblock In \emph{Proceedings of the IEEE/CVF International Conference on Computer Vision}, pages 7777--7787, 2023.

\bibitem[Shamsian et~al.(2020)Shamsian, Kleinfeld, Globerson, and Chechik]{shamsian2020learning}
Aviv Shamsian, Ofri Kleinfeld, Amir Globerson, and Gal Chechik.
\newblock Learning object permanence from video, 2020.

\bibitem[Shan et~al.(2020)Shan, Geng, Shu, and Fouhey]{Shan2020understanding}
Dandan Shan, Jiaqi Geng, Michelle Shu, and David Fouhey.
\newblock Understanding human hands in contact at internet scale.
\newblock In \emph{CVPR}, 2020.

\bibitem[Tenenbaum et~al.(2011)Tenenbaum, Kemp, Griffiths, and Goodman]{tenenbaum2011how}
Joshua~B. Tenenbaum, Charles Kemp, Thomas~L. Griffiths, and Noah~D. Goodman.
\newblock How to grow a mind: Statistics, structure, and abstraction.
\newblock \emph{Science}, 331\penalty0 (6022):\penalty0 1279--1285, 2011.

\bibitem[van~den Oord et~al.(2019)van~den Oord, Li, and Vinyals]{oord2019representation}
Aaron van~den Oord, Yazhe Li, and Oriol Vinyals.
\newblock Representation learning with contrastive predictive coding, 2019.

\bibitem[Vasudevan et~al.(2020)Vasudevan, Dai, and Van~Gool]{vasudevan2020semantic}
Arun~Balajee Vasudevan, Dengxin Dai, and Luc Van~Gool.
\newblock Semantic object prediction and spatial sound super-resolution with binaural sounds.
\newblock In \emph{European conference on computer vision}, pages 638--655. Springer, 2020.

\bibitem[Vasudevan et~al.(2022)Vasudevan, Dai, and Van~Gool]{vasudevan2022sound}
Arun~Balajee Vasudevan, Dengxin Dai, and Luc Van~Gool.
\newblock Sound and visual representation learning with multiple pretraining tasks.
\newblock In \emph{Proceedings of the IEEE/CVF Conference on Computer Vision and Pattern Recognition}, pages 14616--14626, 2022.

\bibitem[Wang et~al.(2022)Wang, Yue, Xu, Hassani, Kulikov, Orlov, Song, Shi, and Huang]{wang2022adafocusv3}
Yulin Wang, Yang Yue, Xinhong Xu, Ali Hassani, Victor Kulikov, Nikita Orlov, Shiji Song, Humphrey Shi, and Gao Huang.
\newblock Adafocusv3: On unified spatial-temporal dynamic video recognition, 2022.

\bibitem[Wu et~al.(2023)Wu, Chen, Zhang, Hui, Berg-Kirkpatrick, and Dubnov]{wu2023large}
Yusong Wu, Ke Chen, Tianyu Zhang, Yuchen Hui, Taylor Berg-Kirkpatrick, and Shlomo Dubnov.
\newblock Large-scale contrastive language-audio pretraining with feature fusion and keyword-to-caption augmentation.
\newblock In \emph{ICASSP 2023-2023 IEEE International Conference on Acoustics, Speech and Signal Processing (ICASSP)}, pages 1--5. IEEE, 2023.

\bibitem[Yang et~al.(2024)Yang, Grady, Brahmbhatt, Vasudevan, Kemp, and Hays]{yang2024kidnappable}
Mengyu Yang, Patrick Grady, Samarth Brahmbhatt, Arun~Balajee Vasudevan, Charles~C Kemp, and James Hays.
\newblock The un-kidnappable robot: Acoustic localization of sneaking people.
\newblock In \emph{2024 IEEE International Conference on Robotics and Automation (ICRA)}, pages 985--992. IEEE, 2024.

\bibitem[Yang et~al.(2025)Yang, Chen, Pei, Agarwal, Vasudevan, and Hays]{yang2025clink}
Mengyu Yang, Yiming Chen, Haozheng Pei, Siddhant Agarwal, Arun~Balajee Vasudevan, and James Hays.
\newblock Clink! chop! thud! -- learning object sounds from real-world interactions.
\newblock In \emph{ICCV}, 2025.

\bibitem[Zhang et~al.(2023)Zhang, Gong, Zhang, Li, Qiao, Ouyang, and Yue]{zhang2023meta}
Yiyuan Zhang, Kaixiong Gong, Kaipeng Zhang, Hongsheng Li, Yu Qiao, Wanli Ouyang, and Xiangyu Yue.
\newblock Meta-transformer: A unified framework for multimodal learning.
\newblock \emph{arXiv preprint arXiv:2307.10802}, 2023.

\bibitem[Zhou et~al.(2023{\natexlab{a}})Zhou, Arnab, Sun, and Schmid]{zhou2023can}
Xingyi Zhou, Anurag Arnab, Chen Sun, and Cordelia Schmid.
\newblock How can objects help action recognition?
\newblock In \emph{Proceedings of the IEEE/CVF Conference on Computer Vision and Pattern Recognition}, pages 2353--2362, 2023{\natexlab{a}}.

\bibitem[Zhou et~al.(2023{\natexlab{b}})Zhou, Arnab, Sun, and Schmid]{zhou2023objects}
Xingyi Zhou, Anurag Arnab, Chen Sun, and Cordelia Schmid.
\newblock How can objects help action recognition?, 2023{\natexlab{b}}.

\bibitem[Zhu et~al.(2023)Zhu, Lin, Ning, Yan, Cui, HongFa, Pang, Jiang, Zhang, Li, Zhang, Li, Liu, and Yuan]{zhu2023languagebind}
Bin Zhu, Bin Lin, Munan Ning, Yang Yan, Jiaxi Cui, Wang HongFa, Yatian Pang, Wenhao Jiang, Junwu Zhang, Zongwei Li, Cai~Wan Zhang, Zhifeng Li, Wei Liu, and Li Yuan.
\newblock Languagebind: Extending video-language pretraining to n-modality by language-based semantic alignment, 2023.

\end{thebibliography}
